\documentclass[10pt,twocolumn,letterpaper]{article}

\usepackage{cvpr}
\usepackage{times}
\usepackage{epsfig}
\usepackage{graphicx}
\usepackage{amsmath}
\usepackage{amssymb}
\usepackage{color}
\usepackage{floatrow}
\newfloatcommand{capbtabbox}{table}[][\FBwidth]
\usepackage[pagebackref=true,breaklinks=true,letterpaper=true,colorlinks,bookmarks=false]{hyperref}

 \cvprfinalcopy 

\ifcvprfinal\fi
\begin{document}

\title{Learning Better Features for Face Detection with Feature Fusion and Segmentation Supervision}

\author{Wanxin Tian$^{1,2}$, Zixuan Wang$^{1}$, Haifeng Shen$^{2}$, Weihong Deng$^{1}$,Yiping Meng$^{2}$,\\ Binghui Chen$^{1}$, Xiubao Zhang$^{2}$,Yuan Zhao$^{2}$, Xiehe Huang$^{1,2}$\\
$^1$Beijing University of Posts and Telecommunications, Beijing 100876, China\\
$^2$AI Labs, Didi Chuxing, Beijing  100193, China\\
{\tt\small \{tianwanxin,princexuan,whdeng,chenbinghui,xiehe.huang\}@bupt.edu.cn}\\
{\tt\small\{shenhaifeng,mengyipingkitty,zhangxiubao,zhaoyuanjason\}@didiglobal.com} 
}

\maketitle

\begin{abstract}
The performance of face detectors has been largely improved with the development of convolutional neural network. 
However, it remains challenging for face detectors to detect tiny, occluded or blurry faces. 
Besides, most face detectors can’t locate face’s position precisely and can’t achieve high Intersection-over-Union (IoU) scores.
 We assume that problems inside are inadequate use of supervision information and imbalance between semantics and details at all level feature maps in CNN even with Feature Pyramid Networks (FPN)~\cite{DBLP:journals/corr/LinDGHHB16}. 
  
 In this paper, we present a novel single-shot face detection network, named DF$^2$S$^2$ (Detection with Feature Fusion and Segmentation Supervision), 
 which introduces a more effective feature fusion pyramid and a more efficient segmentation branch on ResNet-50~\cite{DBLP:journals/corr/HeZRS15} to handle mentioned problems. 
 Specifically, inspired by FPN and SENet~\cite{DBLP:journals/corr/abs-1709-01507}, 
 we apply semantic information from higher-level feature maps as contextual cues to augment low-level feature maps via a spatial and channel-wise attention style, 
 preventing details from being covered by too much semantics and making semantics and details complement each other. 
 
 We further propose a semantic segmentation branch to best utilize detection supervision information meanwhile applying attention mechanism in a self-supervised manner.
 The segmentation branch is supervised by weak segmentation ground-truth (no extra annotation is required) in a hierarchical manner, deprecated in the inference time so it wouldn’t compromise the inference speed. 
 We evaluate our model on WIDER FACE~\cite{DBLP:journals/corr/YangLLT15b}  dataset and achieved state-of-art results.

\end{abstract}
\section{Introduction}

\begin{figure*}
\begin{center}
   \includegraphics[width=0.8\linewidth]{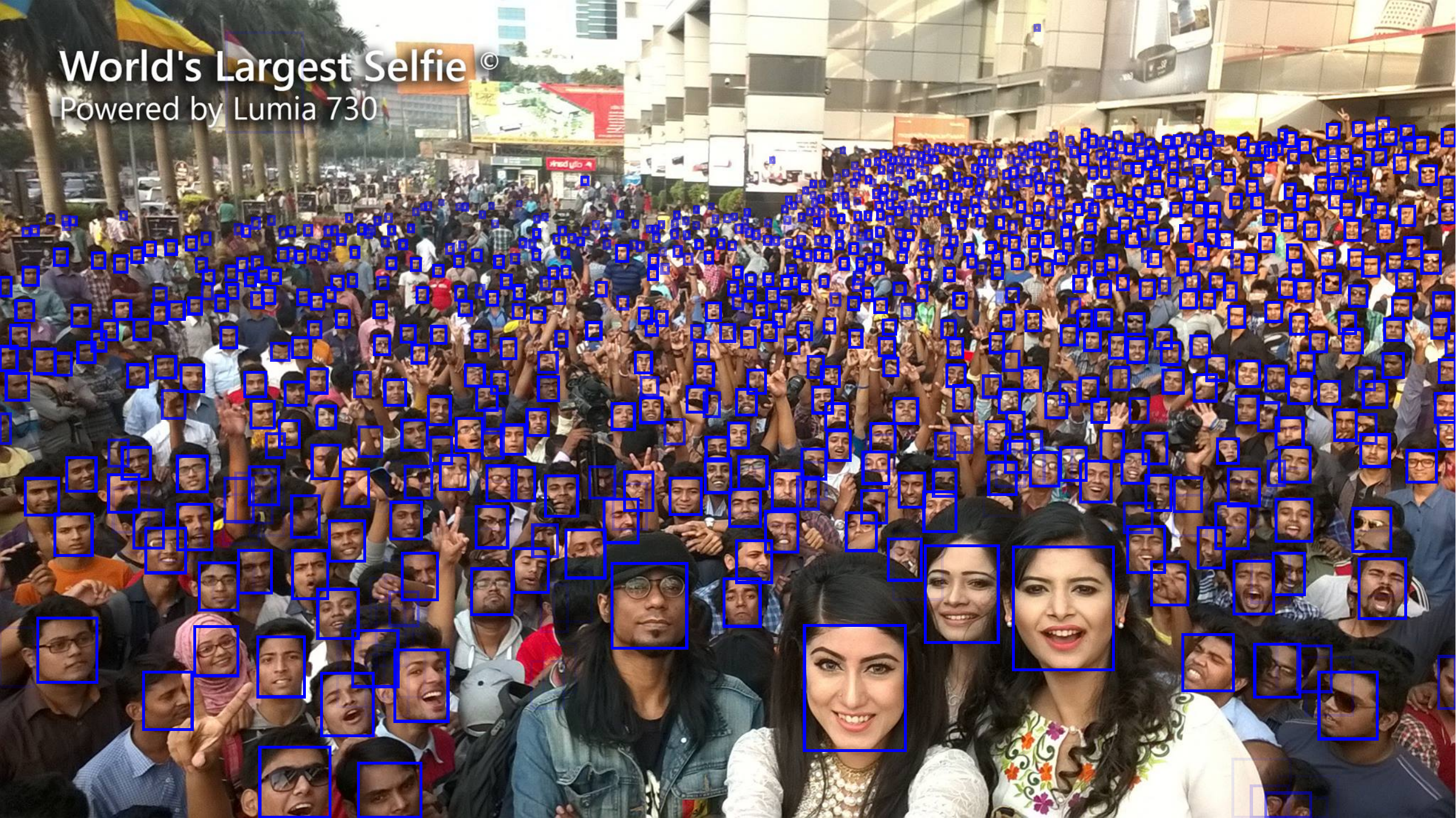}
\end{center}
   \caption{An example of face detection with our proposed DF$^2$S$^2$. In the above image, our model can find 827 faces out of 1000 facial images present with little false positives. The detection confidence scores are positively correlative with transparency of bounding-boxes. Best viewed in color.}
\label{fig:selfie}
\end{figure*}

Face detection is an essential step for many subsequent face-related applications, 
such as face alignment~\cite{DBLP:journals/corr/ZhuLLSL15}  face recognition~\cite{DBLP:conf/cvpr/ZhuLYYL15}  and face verification~\cite{DBLP:conf/cvpr/TaigmanYRW14}, \etc. 
It has been well developed over the past few decades.
 Following the pioneering work of Viola-Jones face detector~\cite{DBLP:conf/iccv/ViolaJ01}, most of early works focused on crafting effective features manually and training powerful classifiers. 
 But these hand-crafted features are indiscriminative and each component is isolated, making the face detection pipeline sub-optimal.

 Recently, object detection borrows ImageNet~\cite{DBLP:conf/nips/KrizhevskySH12} pre-trained models as the backbone from image classification and have acquired significant improvements. 
For the task of image classification only needs semantics to recognize the category, feature maps own more semantic information and less detailed information with going deeper in CNN,
However, both of semantics and details are in demand for face detectors to detect faces in different locations with various scales and charicteristics.
 Consequently FPN~\cite{DBLP:journals/corr/LinDGHHB16} presents a divide and conquer principle that different scales of objects are collected and distributed to different feature layers,
 with a top-down architecture attached to maintain both the high spatial resolution and semantic information.
 
 We observe that FPN obtains semantic enrichment at lower-level layers by adding deformation of higher-level feature maps to lower-level feature maps,
 which may cause too much semantics from higher-level feature maps damages details in lower-level feature maps.
As can be seen in~\cite{DBLP:journals/corr/ZeilerF13},  semantics represents the more semantic meaningful patterns whose receptive filed is larger,
while details represent basic visual patterns whose receptive filed is smaller.
 Intuitively, they will make conflicts when fusing semantics and details in an addition manner.
 So, the key of feature fusion is to prevent conflicts among different feature maps and loss of information in the process of transformation.
To obtain semantic enrichment at lower-level layers and meantime prevent details from being covered by too much semantics,
we propose a novel feature pyramidal structure to fuse higher-level feature maps and lower-level feature maps in a spatial and channel-wise attention manner.
More specially, we apply semantic information of higher-level feature maps as contextual cues to element-wisely multiply lower-level feature maps.
We further avoid loss of semantic information by applying transposed convolution (also called deconvolution~\cite{DBLP:conf/cvpr/ZeilerKTF10}) to transform feature maps.

Secondly, most works divide the task of detection into the classification task and the regression task, both of which handle pre-set anchors.
When anchors match objects not well, the objects would be ignored with a waste of detection supervision information, making optimization sub-optimal.
 So anchor assign strategy decides the ceiling of performance of anchor-based face detection. 

In this paper, to complement anchor assign strategy and best utilize detection supervision information, 
we introduce an efficient segmentation branch like~\cite{DBLP:journals/corr/LongSD14}. 
The segmentation branch is trained with bounding-box segmentation ground-truth in a hierarchical manner. 
The segmentation branch can help networks learn more discriminative features from object regions, which has been proved helpful in~\cite{Song_2018_CVPR},  in a self-supervised manner.
We employ the segmentation in the training phase to apply attention mechanism 
-- a dynamic feature extractor that combines contextual fixations over times, as CNN features are naturally spatial, channel-wise and multi-layer~\cite{DBLP:conf/cvpr/ChenZXNSLC17}, 
and there will be no extra parameters in the inference time.
 We conduct extensive experiments on WIDER FACE~\cite{DBLP:journals/corr/YangLLT15b} benchmarks to validate the efficacy of our proposed structure.

As a summary, the main contributions of this paper include the following:
\begin{itemize}
\item We propose a novel feature pyramidal structure to apply semantic information in higher-level feature maps as contextual cues 
to augment semantics in lower-level feature maps in a spatial and channel-wise attention manner.
\item We improve the typical deep single shot detectors by making up for anchor mechanism 
with a semantic segmentation branch to apply attention mechanism, without compromising the inference speed.
\item We present a novel single-shot face detector, called DF$^2$S$^2$ (Detection with Feature Fusion and Segmentation Supervision),
 which can learn better features for face detection and therefore can address well the occlusion and multi-scale issues.
 We demonstrate a qualitative result of our DF$^2$S$^2$ in Figure \ref{fig:selfie}.
 \item Extensive experiments are carried out on WIDER FACE dataset to demonstrate the efficiency and effectiveness of our model.
\item We achieve state-of-art results on WIDER FACE dataset with real-time inference speed.
\end{itemize}

\begin{figure*}
\begin{center}
   \includegraphics[width=0.8\linewidth]{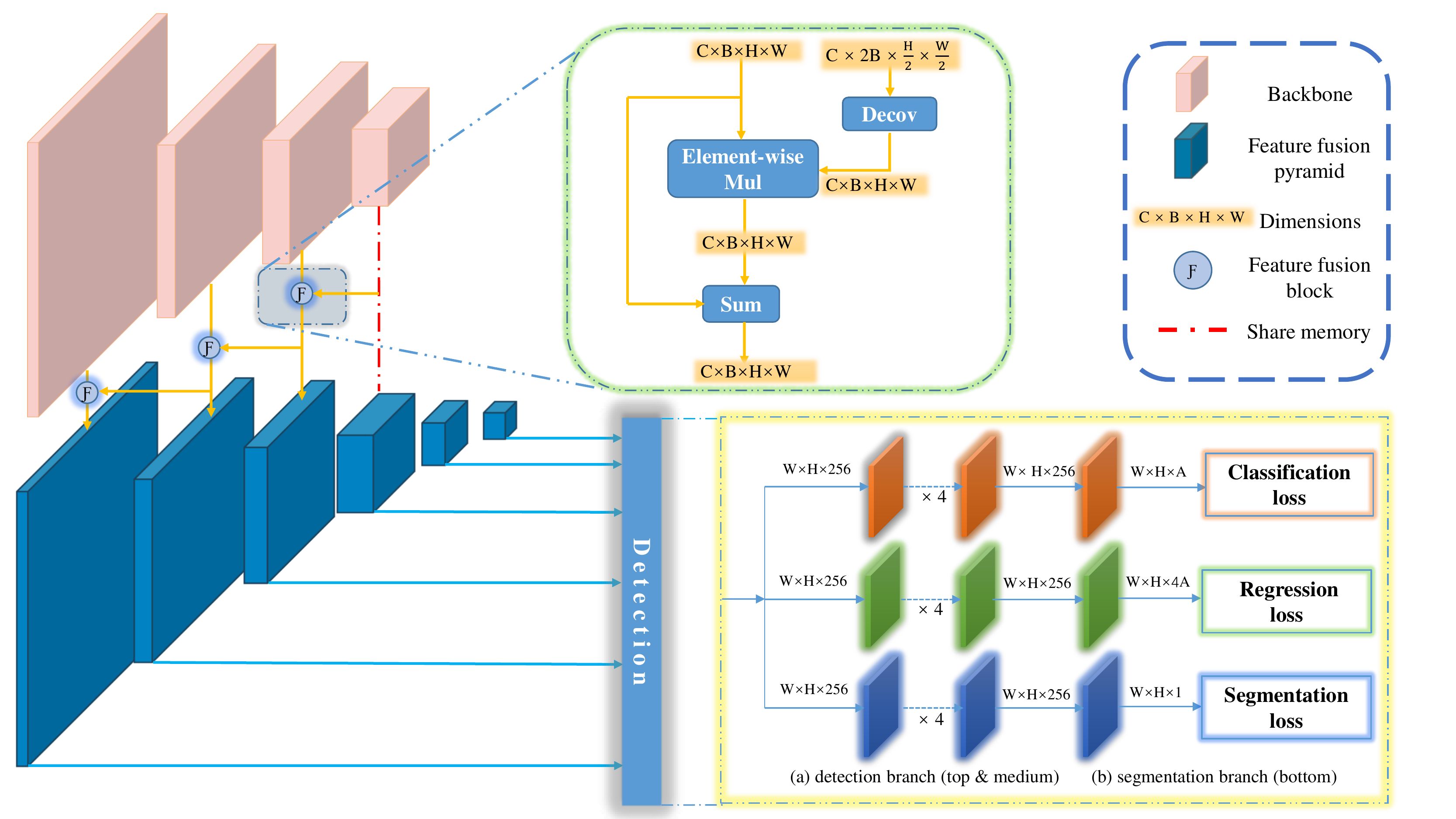}
\end{center}
   \caption{The network architecture of our proposed DF$^2$S$^2$. It consists of backbone network, feature fusion pyramid structure and the detection module. The detection module contains the detection branch and the segmentation branch.}
\label{fig:overat}
\end{figure*}

\section{Related work}

{\bf Face Detection.} 
Benefiting from the remarkable achievement of deep convolutional networks on image classification~\cite{DBLP:conf/nips/KrizhevskySH12}  
and object detection~\cite{DBLP:journals/corr/RedmonDGF15,DBLP:journals/corr/LiuAESR15,DBLP:journals/corr/RenHG015}, 
CNN-based face detectors have also gained much performance improvement recently. 
Deep learning models trained on large-scale image datasets provide more discriminative features for face detector compared to traditional hand-crafted features. 
Besides, the end-to-end training style promotes better optimization. 
The performance gap between human and artificial face detectors has been significantly closed.
Based on whether following the proposal and refine strategy, 
deep learning methods can be divided into two categories:  one-stage approaches, such as 
YOLO~\cite{DBLP:journals/corr/RedmonDGF15}, 
SSD~\cite{DBLP:journals/corr/LiuAESR15}, RetinaNet~\cite{DBLP:journals/corr/abs-1708-02002}, RefineDet~\cite{DBLP:conf/cvpr/ZhangWBLL18},
and two-stage approaches such as Faster R-CNN~\cite{DBLP:journals/corr/RenHG015}, R-FCN~\cite{DBLP:journals/corr/DaiLHS16}. 
UnitBox~\cite{DBLP:journals/corr/YuJWCH16}  presents a new intersection-over-union (IoU) loss to directly optimize IOU target. 
HR~\cite{DBLP:journals/corr/HuR16}  builds multi-level image pyramids for multi-scale training and testing which finds upscaled tiny faces. 
DCFPN~\cite{DBLP:journals/ijon/ZhangZLWSL18} and FaceBoxes~\cite{DBLP:conf/icb/ZhangZLSWL17} design a lightweight face detection network based on SSD to achieve CPU real-time speed with promising result.
S$^3$FD~\cite{DBLP:journals/corr/abs-1708-05237}  and SFDet~\cite{zhang2019single-shot}  addresses this with scale-equitable framework and new anchor matching strategy. 
RetinaNet~\cite{DBLP:journals/corr/abs-1708-02002} introduces a new focal loss to relieve the class imbalance problem. 
PyramidBox~\cite{DBLP:journals/corr/abs-1803-07737} utilizes contextual information with improved SSD network structure. 

{\bf Attention Mechanism.} 
Attention mechanism has been proved effective in various computer vision tasks such as image captioning~\cite {DBLP:journals/corr/XuBKCCSZB15} and visual question answering~\cite{DBLP:journals/corr/ChenWCGXN15}. 
It is inspired by the reasonable assumption that human
vision does not tend to process a whole image in its
entirety at once; instead, one only focuses on selective parts
of the whole visual space when and where as needed~\cite{Corbetta2002}.
Specifically, rather than encoding an image into a static vector,
attention mechanism allows image features to evolve from the sentence context at hand, 
resulting in richer and longer descriptions for cluttered images~\cite{DBLP:journals/corr/ChenZXNSC16}. 
In this way, attention mechanism
can be considered as a dynamic feature extraction mechanism
that combines contextual fixations over times~\cite{DBLP:journals/corr/MnihHGK14}.

{\bf Segmentation branch.}  
Segmentation branch is initially used in the semantic segmentation task~\cite {DBLP:journals/corr/LongSD14} to classify each pixel in one image. 
However, Papandreou \etal~\cite{DBLP:conf/iccv/GidarisK15} proved that weakly annotated data such as
bounding-boxes and image-level labels can also be utilized for semantic segmentation.
He  \etal~\cite{DBLP:journals/corr/HeGDG17}  showed that multi-task training of object detection and instance segmentation 
can help to improve the object detection task with extra instance segmentation annotation.
However, we do not consider extra annotation in our work. 
Dense-Box~\cite{DBLP:journals/corr/HuangYDY15}  utilizes a unified end-to-end fully convolutional network to detect confidence and bounding box directly. 
FAN~\cite{DBLP:journals/corr/abs-1711-07246}  proposes an anchor-level attention into RetinaNet to detect the occluded faces. 
In this paper, we introduce segmentation branch into the popular single shot detector with weak segmentation ground-truth, 
applying attention mechanism without compromising inference speed. 

{\bf Feature Pyramid.} 
Feature pyramid is a structure which applies skip-connection to combine semantic meaningful features with semantically weak but visually strong features.
FPN~\cite{DBLP:journals/corr/LinDGHHB16} proposed a top-down architecture to use high-level semantic feature maps at all scales.
FANet~\cite{DBLP:journals/corr/abs-1712-00721} agglomerates multi-scale features to augment lower-level feature maps in a concatenation style.
In this paper, we propose a novel feature fusion connection which aggregate multi-scale features in a spatial and channel-wise attention manner.

\section{Detection with Feature Fusion and Segmentation Supervision (DF$^2$S$^2$ )}
In this section, we present our DF$^2$S$^2$ framework for face detection. 
First, we present the overall architecture in Section \ref{sec:overat}. 
Then we propose a novel feature fusion pyramid structure replacing FPN (Feature Pyramid Networks) in Section \ref{sec:seg}
 and a segmentation branch to balance semantics and details at all level detection feature maps in Section \ref{sec:ffp}. 
Finally, we will introduce the associated training methodology in Section \ref{sec:train}.

\subsection{Overall architecture}

\newcommand{\tabincell}[2]{\begin{tabular}{@{}#1@{}}#2\end{tabular}}

\label{sec:overat}

Our goal is to learn more discriminative hierarchical features with enriched semantics and details at all levels to detect hard faces, like tiny faces, partially occluded faces, etc. 
Figure \ref{fig:overat} illustrates our proposed network with feature fusion pyramid  and the segmentation branch. 
To obtain strong capability of generality, we consider the widely used ResNet-50 as the backbone CNN architecture and mimic S$^3$FD~\cite{DBLP:journals/corr/abs-1708-05237}  to build our single-shot multi-scale face detector. 

First, we build our feature fusion pyramid structure based on four layers of $\{res2/_2, res3/_3, res4/_5, res5/_2\}$ from ResNet-50 (colored white in the left-upper part of Figure \ref{fig:overat}).
The structure takes four feature maps from these layers as inputs,  
and generates  four corresponding new feature maps with augmented semantics and details of $\{FFP_2, FFP_3, FFP_4, FFP_5\}$ (highlighted as blue feature maps in the left-bottom part of Figure \ref{fig:overat}), 
 whose spatial resolution and the number of channels are identical to input feature maps, respectively. 
 To get larger receptive field to detect bigger faces, we simply max-pool the $FFP_5$ feature map twice in succession to get extra two feature maps of $\{FFP_6, FFP_7\}$. 
 The six detection feature maps have strides of $\{4, 8, 16, 32, 64, 128\}$, respectively.
 As shown in Figure \ref{fig:overat}, the detection and segmentation is performed on feature maps of $FFP_n$ (ranging from $2$ to $7$) layers.

In the detection branch, the classification subnet applies four $3 \times 3$ convolution layers each with 256 filters, 
followed by a $3 \times 3$ convolution layer with $K \times A$ filters where $K$ means the number of classes and $A$ means the number of anchors per location. 
For face detection $K = 1$ since we use sigmoid activation, and we use $A = 6$ in most experiments. 
All convolution layers in this subnet share parameters across all pyramid levels to accelerate convergence of parameters. 
The regression subnet is identical to the classification subnet except that it terminates in $4 \times {A}$ convolution filters with linear activation.

To enhance the correlation between the classification subnet and the regression subnet and improve the separation of semantic supervision information and location supervision information,
 the parameters of the convolutional layers are shared across the detection branch, except for the last prediction layer.

\subsection{Segmentation branch}

\label{sec:seg}

To make up for anchor assign strategy and make full use of detection supervision information,
we present our effective and efficient segmentation branch.
As is shown in the right lower part of Figure \ref{fig:overat}, the segmentation branch is parallel to the classification subnet and the regression subnet in the head-architecture.
It takes feature maps of {$FFP_2, FFP_3, FFP_4, FFP_5, FFP_6, FFP_7$} as inputs, the same with the detection branch,
and is supervised with the bounding-box level segmentation ground-truth in a hierarchical manner.
Following the match principle of~\cite{DBLP:journals/corr/LuoLUZ17} and S$^3$FD~\cite{DBLP:journals/corr/abs-1708-05237},
 these hierarchical segmentation maps are associated to the ground-truth faces matching their corresponding receptive field.
The receptive field is identical between the segmentation branch and the detection branch to make sure they focus on the same range of face scales.
Consequently, our segmentation helps networks learn more discriminative features from face regions, 
and further makes the tasks of classification and regression easier for detection branch, promoting better optimization.
  
 We add  four $3 \times 3$ convolutional layers each with $256$ filters after input feature maps,
 followed by one $3 \times 3$ convolutional layer with $K$ filters where $K$ means the number of classes. 
 For face detection $K = 1$ since we use sigmoid activation. 
To enhance the impacts of segmentation supervision information to the detection branch and preserve more parameters of segmentation branch,
parameters of the former four convolutional layers are further shared with the detection branch.
The segmentation branch is deprecated in the inference time for uselessness of segmentation prediction maps. 

Our advantage over other usings of segmentation branch is that,
instead applying segmentation prediction maps (like FAN~\cite{DBLP:journals/corr/abs-1711-07246}) or the intermediate result (like DES~\cite{DBLP:journals/corr/abs-1712-00433}) to activate feature maps of main branch,
we apply the attention mechanism in a self-supervised manner without extra parameters and activation operation.
Besides,  there is little redundant background region in the bounding-box segmentation ground-truth for face detection, as face regions usually take most places of the bound-box ground-truth,
 when chaotic backgrounds interfere the learning of discriminative features from object regions.
 Mathematically, the average IoU (Intersection of Union) between actual segmentation ground truth and bounding-box ground truth for face is so high that influence of redundant context regions is negligible.
 \label{sec:fan}

\setlength{\tabcolsep}{4pt}
\begin{table*}
\begin{center}
\caption{The comparative experiments with RetinaNet (Baseline) and FAN~\cite{DBLP:journals/corr/abs-1711-07246} on the WIDER FACE validation set. Minimum size of input images for FAN is 1000.}
\label{table:mask}
\begin{tabular}{c|cccc|ccc}
\hline\hline
BaseNet&Attention&\tabincell{c}{Data augmention}&\tabincell{c}{Multi-scale}&\tabincell{c}{Segmentation branch}&\tabincell{c}{AP (easy)}&\tabincell{c}{AP (medium)}&\tabincell{c}{AP(hard)}\\

\hline
\hline
RetinaNet & & & & & 92.9&90.9 &74.2 \\
\hline
FAN  & \checkmark & & & & 88.4&88.4 &80.9 \\
FAN & \checkmark& \checkmark & & &94.0 &93.0 &86.8 \\
FAN & \checkmark& & \checkmark& &91.7 &90.4 &84.2 \\
FAN & \checkmark & \checkmark & \checkmark & & 95.3& 94.2&88.8 \\
\hline
Ours & & & & \checkmark &92.9 &91.4 &81.4\\
Ours & & \checkmark & & \checkmark & 94.0&92.8 &84.0 \\
Ours & & & \checkmark& \checkmark & 94.1&92.8 &86.8 \\
Ours & & \checkmark & \checkmark& \checkmark &95.2 &94.3 & 88.8\\
\hline
\end{tabular}
\end{center}
\end{table*}

\subsection{Feature fusion pyramids}

\label{sec:ffp}

Figure \ref{fig:overat} illustrates the idea of the proposed feature fusion pyramid and feature fusion block (called ``$F$-block" for short).
We apply the ``$F$-block'' to fuse different feature maps from top to bottom recursively. 
Mathematically, we express our feature fusion method as $\phi_{i} =   F(\phi_{i+1},\phi_{i};\theta)$ and detail our $F$ as following formula:
\setlength\abovedisplayskip{4pt}
\setlength\belowdisplayskip{4pt}
\begin{equation} \label{eq:ep1} 
                  \phi_{i} =   \phi_{i} \cdot \Psi(\phi_{i+1};\theta)  +   \phi_{i}
\end{equation}
Where  $\phi_{i}$ and $\phi_{i+1}$  represent the shallower feature map and the deeper one respectively. 
 $\Psi$ represents the transposed convolution operation on the high-level feature map,
 $\theta$ represents the parameters of the transposed convolution.
 $\phi_{i}$ on the left side of the formula represents the new generated feature map after fusion
 and would continue to participate in the process of feature fusion with lower-level feature maps until the lowest. 
The element-wise multiplication (represented as $\cdot$) 
can be seen as the combination of the spatial and channel-wise attention that maximize mutual information between lower-level and higher-level representations.
 Furthermore, in order to enhance the detailed information which is essential for detecting hard faces,
the low-level feature map is then added to the previously generated feature map after element-wise multiplication.

It is worth noting that when doing transformation to the higher-level feature maps, 
we apply transposed convolution instead of the combination of up-sampling operation and one convolution.
 On one hand, if we first up-sample the high-level feature map, it will double the number of parameters for the following convolutional operation, 
 which will compromise the inference speed. On the other hand, if we first convolute the high-level feature map to half the number of channels, 
 we may lose some of the semantics of high-level feature map inevitably, hurting the fusion of features. 
 So we take advantages of the transposed convolution, changes the spatial resolution and channels of feature map in one step.

\subsection{Training}  \label{sec:train}

In this section, we introduce our anchor assign strategy, loss function, data augmentation and other implementation details.

{\bf Anchor assign strategy.} Following the scales designing for anchors in S$^3$FD, we have six detector layers each associated with a specific scale anchor. 
Specifically, scales of anchors are carefully designed according to effective receptive field, making the size of anchors four times as the stride of each layer. 
Thus, we set our anchors from area of $16^2$ to $512^2$ on pyramid levels. 
In addition, the aspect ratio for our anchor is set as $1$ and $1.5$, because most of frontal faces are approximately square and profile faces can be considered as a $1:1.5$ rectangle. 
Specifically, anchors are assigned to a ground-truth box with the highest IoU larger than $0.5$, and to background if the highest IoU is less than $0.4$. Unassigned anchors are ignored during training.

{\bf Loss function.} In the training phase, an extra cross-entropy loss function for the segmentation branch will be added in conjunction with the original face detection loss function to jointly optimize model parameters:
\label{sec:loss}
\begin{align} \label{eq:loss} 
		L = & \sum_{k} \frac{1}{N_{k}^{c}}\sum_{i \in A_{k} }L_{c}(p_{i},p_{i}^{\ast}) +    \nonumber \\	
                & \lambda_{1} \sum_{k}    \frac{1}{N_{k}^{r}}\sum_{i \in A_{k} } I(p_{i}^{\ast}) L_{r}(t_{i},t_{i}^{\ast}) + \\                  
                 &   \lambda_{2}  \sum_{k}L_{s}(m_{k},m_{k}^{\ast}) \nonumber 
\end{align}

where $k$ is the index of feature fusion pyramids level $(k \in [2, 7])$, and $A_k$ represents the set of anchors defined in pyramid level $P_k$. 
The ground-truth label $p_{i}^{\ast}$ is $1$ if the anchor is positive, $0$ otherwise. 
$p_{i}$   is the predicted classification result from our model. $t_{i}$ is a vector representing the $4$ parameterized coordinates of the predicted bounding box, 
and $t_{i}^{\ast}$ is that of the ground-truth box associated with a positive anchor.

The classification loss $L_{c}(p_{i},p_{i}^{\ast})$ is focal loss introduced in~\cite{DBLP:journals/corr/abs-1708-02002} over two classes (face and background). 
$N_{k}^{c}$ is the number of anchors in $P_k$ which participate in the classification loss computation. 
The regression loss $L_{r}(t_{i},t_{i}^{\ast})$ is smooth $L1$ loss. $ I(p_{i}^{\ast})$ is the indicator function that limits the regression loss only focusing on the positively assigned anchors, and $N_{k}^{r} = \sum_{i \in A_{k}} I(p_{i}^{\ast})$. 
The segmentation loss $L_{s}(m_{i},m_{i}^{\ast})$ is pixel-wise sigmoid cross entropy. 
$m_k$ is the segmentation prediction map generated per level, and $m_{i}^{\ast}$ is the weak segmentation ground-truth described in Section \ref{sec:seg}. 
$\lambda_1$ and $\lambda_2$ are used to balance these three loss terms, here we simply set $\lambda_1 = $1 and discuss more about $\lambda_2$ in Section \ref{sec:abloss}. 

{\bf Data augmentation.} According to the statistics from the WiderFace dataset, there are around $26\%$ of faces with occlusion. 
Among them, around $16\%$ is of serious occlusion. As we are targeting to solve the occluded faces, the number of training samples with occlusion may not be sufficient. 
Thus, we employ random crop data augmentation. 
The performance improvement is significant. 
Besides from the benefits for the occluded face, our random crop augmentation potentially improves the performance of small faces as more small faces will be enlarged after augmentation.\label{sec:aug}

{\bf Other implementation details.} The training starts from fine-tuning ResNet-50 backbone network using SGD with momentum of $0.9$, weight decay of $0.0001$, and a total batch size of $32$ on $4$ GPUs. 
The newly added layers are initialized with “xavier”. 
We train our model for $120$ epochs and a learning rate of $4 \times 10{-3}$ for first 80 epochs and continue training for $20$ epochs with $4 \times 10^{-4}$ and $4 \times 10^{-5}$ . 
Our implementation is based on Detectron~\cite{Detectron2018}, and our source code will be made publicly available.

\begin{figure*}
\begin{center}
   \includegraphics[width=1\linewidth]{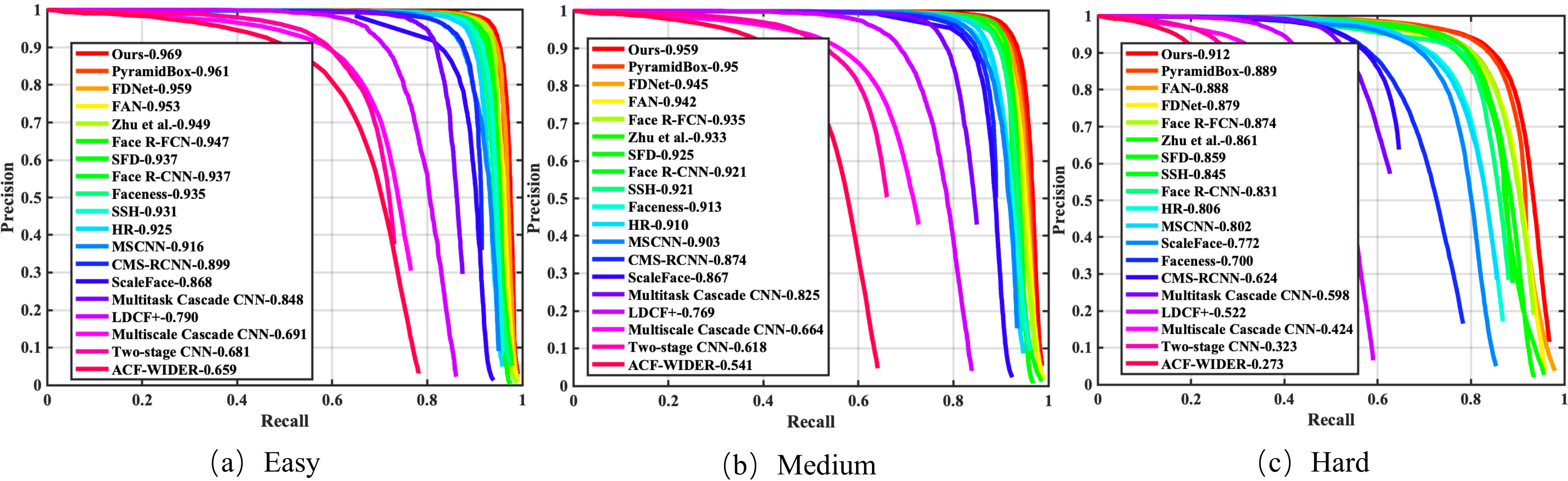}
\end{center}
   \caption{Precision-recall curves on WIDER FACE validation set.}
\label{fig:val}
\end{figure*}

\begin{figure*}
\begin{center}
   \includegraphics[width=1\linewidth]{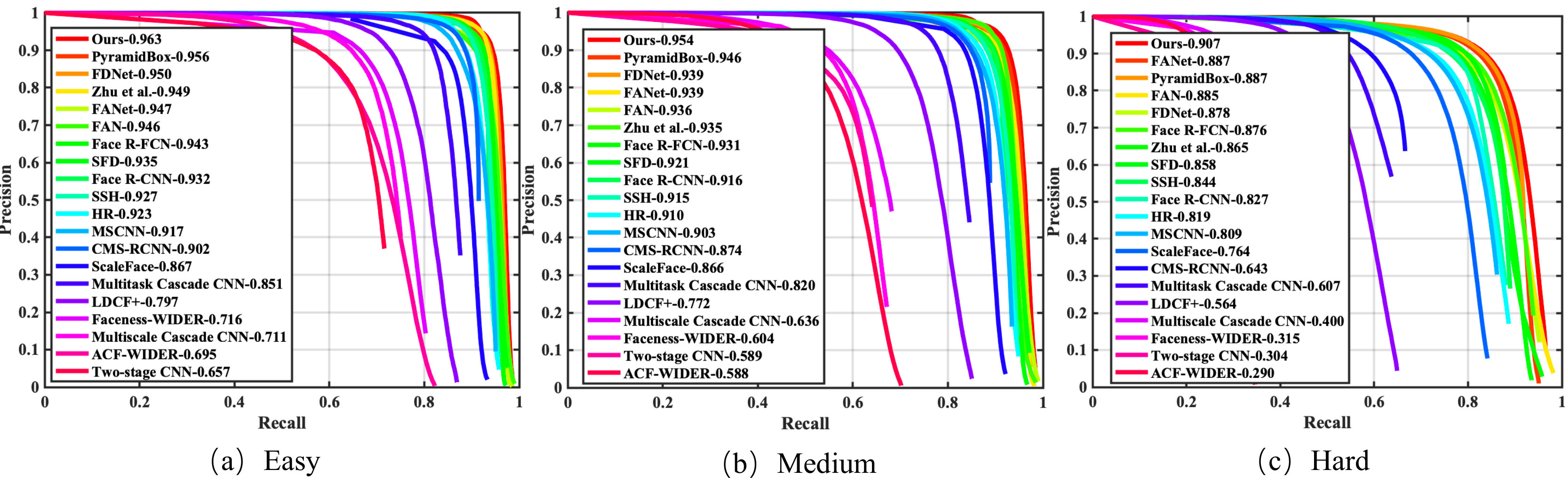}
\end{center}
   \caption{Precision-recall curves on WIDER FACE testing set.}
\label{fig:test}
\end{figure*}

\section{Experiments}

In this section, we first analyze the effectiveness of our segmentation branch and feature fusion pyramids structure on extensive experiments and ablation studies. 
Then, we compare our proposed face detector with the state-of-the-art face detectors on popular face detection benchmarks and finally evaluate the inference speed of the proposed face detector.

\setlength{\tabcolsep}{6pt}
\begin{table}
\begin{center}
\caption{The ablation study of feature fusion pyramid and segmentation branch. 
Seg. stands for the segmentation branch, Aug. stands for the data augmentation method mentioned in Section \ref{sec:aug} and Fus. stands for the feature fusion pyramid.}
\label{table:fus}
\begin{tabular}{c|c|c|c}
\hline
BaseNet&\tabincell{c}{AP\\(easy)}&\tabincell{c}{AP\\(medium)}&\tabincell{c}{AP\\(hard)}\\

\hline

RetinaNet & 92.9&90.9 &74.2 \\
\hline
\tabincell{c}{Ours\\(Seg.)} &92.9 &91.4 &81.4\\\hline
\tabincell{c}{Ours\\(Seg.$+$Aug.)} & 94.0&92.8 &84.0 \\\hline
\tabincell{c}{Ours\\(Fus.)} & 93.7&91.7 &80.3 \\\hline
\tabincell{c}{Ours\\(Fus.$+$Seg.$+$Aug.)} & 94.2 &93.2 & 85.5 \\ \hline
\end{tabular}
\end{center}
\end{table}
\setlength{\tabcolsep}{1.4pt}
{\bf Datasets.} We conduct model analysis on the WIDER FACE dataset~\cite{DBLP:journals/corr/YangLLT15b},
 which has 32,203 images with about 400k faces for a large range of scales. 
 It consists of three subsets: $40\%$ for training, $10\%$ for validation, and $50\%$ for testing.
  The annotations of training and validation sets are online available. 
  According to the difficulty of detection tasks, it has three splits: Easy, Medium and Hard. 
  The evaluation metric is mean average precision (mAP) with Interception-of-Union (IoU) threshold as $0.5$. 
  We train our model on the training set of WIDER FACE, and evaluate it on the validation and testing set. 
  If not specified, the results in Table \ref{table:mask}, \ref{table:fus} and \ref{table:lambda} are obtained by single scale testing in which the shorter size of image is resized to 800 while keeping image aspect ratio. 

{\bf Baseline.} To evaluate our contributions, we conduct comparative experiments with our baseline. 
We adopt the closely related detector RetinaNet as the baseline. 
RetinaNet achieved the state-of-the-art results on several well-known face detection benchmarks. 
It inherited the standard SSD framework with relieving the class imbalance problem via a novel focal loss function. 
We train all models with identical strategies mentioned in \ref{sec:train} for fair comparison. 

\subsection{Ablation studies on segmentation branch}
To examine the impact of our segmentation branch, we have conducted lots of comparative experiments as can be seen in Table \ref{table:mask}.
The comparison between the first and the sixth rows in Table \ref{table:mask} indicates that our segmentation branch effectively improve the performance, especially for small faces. 
The AP is increased by $0.5\%$ and $7.2\%$ on WIDER FACE medium and hard subsets, respectively, without bells and whistles.
The great advancement on detecting tiny faces demonstrates that our segmentation indeed help the model learn more robust features from small faces and make features highlight face regions.

\setlength{\tabcolsep}{6pt}
\begin{table}
\begin{center}
\caption{Ablation results evaluated on WIDER FACE validation set. $\lambda_{2}$  is the hyper-parameter controlling the trade-off between the segmentation loss and detection loss in Eq. (\ref{eq:loss}).}
\label{table:lambda}
\begin{tabular}{c|c|c|c}
\hline\noalign{\smallskip}
BaseNet&\tabincell{c}{AP\\(easy)}&\tabincell{c}{AP\\(medium)}&\tabincell{c}{AP\\(hard)}\\

\hline
\tabincell{c}{Ours($\lambda_{2}$=0.05)} & 94.2&93.2 &85.5\\\hline
\tabincell{c}{Ours($\lambda_{2}$=0.1)} & 94.1&93.0 &85.4\\\hline
\tabincell{c}{Ours($\lambda_{2}$=0.2)} & 94.3&93.2 &85.3 \\\hline
\tabincell{c}{Ours($\lambda_{2}$=0.5)} & 94.4&93.1 &83.4 \\\hline
\tabincell{c}{Ours($\lambda_{2}$=1.0)} & 94.3&92.9 &83.5\\

\noalign{\smallskip}\hline
\end{tabular}
\end{center}
\end{table}

Besides, we further compare our model with FAN~\cite{DBLP:journals/corr/abs-1711-07246}, which also introduce the segmentation branch to make networks pay more attention on face regions.
Differences between our model and FAN has been analyzed in \ref{sec:fan}.
The results of FAN are obtained by single scale testing in which the shorter size is resized to 1000 while keeping  aspect ratio. 
Without data augmentation and multi-scale testing, our performance is $4.5\%$, $0.5\%$ and $0.5\%$ higher on easy, medium and hard subset respectively. 
It indicates that our segmentation branch can bring more effectiveness with self-learning of models and self-adaption of parameters by more comprehensive supervision.
On the contrary, FAN apply the attention maps to weight  features map in spatial direction for highlighting face features, when taking risks of hurting semantics and details.
When coming to data augmentation and multi-scale training, our model is comparable with FAN.
Our advantage is that there is no extra parameter in inference time with almost competitive improvements.

\subsection{Ablation studies on feature fusion}
We build ``Ours (Fus.)'' replacing the FPN part in RetinaNet with our feature fusion pyramids structure and  build ``Ours (Seg.)'' on RetinaNat with the segmentation branch to conduct comparative experiments.
In Table \ref{table:fus}, compared with the plain RetinaNet, ``Ours (Fus.)'' gains improvement $0.8\%$,$0.8\%$ and $6.1\%$ in easy, medium and hard level respectively, 
which validates the efficacy of feature fusion pyramids for enriching semantics and details in a balanced manner, and demonstrates the superiority of our model over FPN.
With the segmentation branch, the performance further improved, $+0.2\%$,$+0.4\%$ and $+1.5\%$ in easy, medium and hard level, shown in column $3$, $5$.
The great improvements on hard subset proved that our feature fusion method can actually enrich semantics in lower-level feature maps without damaging details.

\setlength{\tabcolsep}{6pt}
\begin{table}
\begin{center}
\caption{Evaluation on WIDER FACE validation set (mAP). The red marked part represents the highest score in the corresponding dataset, and the blue represents the second highest score. We follow the similar training strategy in~\cite{DBLP:journals/corr/abs-1809-02693,DBLP:journals/corr/abs-1901-06651} when training DF$^2$S$^2$$^{\ast}$ with ResNet101.}
\label{table:widerface}
\begin{tabular}{c|c|c|c|c}
\hline
Algorithms&Backbone&Easy&Med.&Hard\\
\hline
MTCNN~\cite{DBLP:journals/corr/ZhangZL016}         &-              &  84.8          &    82.5    & 59.8    \\
LDCF+~\cite{DBLP:journals/corr/Ohn-BarT17}       &-              &     79.0       &     76.9   &  52.2    \\
\hline
CMS-RCNN~\cite{DBLP:journals/corr/ZhuZLS16}    &VGG16        &       89.9     &   87.4     &     62.4 \\
MSCNN~\cite{DBLP:journals/corr/ChenHW016}    &VGG16         &     91.6       &    90.3    &   80.2   \\
Face R-CNN~\cite{DBLP:journals/corr/WangLJW17} &VGG16           &     93.7       & 92.1       &  83.1    \\
SSH~\cite{DBLP:journals/corr/abs-1708-03979}    &VGG16       &        93.1    &      92.1  &    84.5  \\
S$^3$FD~\cite{DBLP:journals/corr/abs-1708-05237}      &VGG16        &       93.7     &   92.5     &  85.9    \\
PyramidBox~\cite{DBLP:journals/corr/abs-1803-07737} &VGG16         &  \textbf{\color{blue}96.1}  &   \textbf{\color{blue}95.0}     &  88.9   \\
FANet~\cite{DBLP:journals/corr/abs-1712-00721} &VGG16         &    95.6          &  94.7     & 89.5  \\
\hline
HR~\cite{DBLP:journals/corr/HuR16}     &ResNet101       &    92.5        &    91.0    &    80.6  \\
Face R-FCN~\cite{DBLP:journals/corr/DaiLHS16}   &ResNet101       &     94.7       &    93.5    &  87.3   \\
Zhu~\cite{DBLP:journals/corr/abs-1802-09058}      &ResNet101       &       94.9     &     93.8   &  86.1    \\
ScaleFace~\cite{DBLP:journals/corr/YangXLT17}  &ResNet50        &    86.8        &   86.7     &   77.2   \\
FAN~\cite{DBLP:journals/corr/abs-1711-07246}   &ResNet50        &   95.3       &   94.2     &   88.8   \\
 \textbf{DF$^2$S$^2$(ours) }  &ResNet50        &  95.6        &  94.7     &   \textbf{\color{blue}89.8}   \\
  \textbf{DF$^2$S$^2$$^{\ast}$(ours) }  &ResNet101        &  \textbf{\color{red}96.9}        &  \textbf{\color{red}95.9}      &   \textbf{\color{red}91.2}   \\
\hline
\end{tabular}
\end{center}
\end{table}

\subsection{Experiments on Balancing the loss}

\label{sec:abloss}

Another ablation study is conducted in the weight of the segmentation loss. 
For the absence of the segmentation branch in the inference time, we assume that it may not be optimal to make these losses numerically consistent.
To find the optimal weight, we train our model with different $\lambda_2$'s, \ie, 0.05, 0.1, 0.2, 0.5 and 1.
In Table \ref{table:lambda}, experiments show that $\lambda_{2}=0.05 $ yields the best performance in total.
The small margin among the four performance indicates that our segmentation branch always improve the models with the inside ability of self-optimization and  brings little risk of hurting the detection performance.

\subsection{Evaluation on WIDER FACE benchmark}
We compare our DF$^2$S$^2$ with the state-of-art detectors, such as  PyramidBox, FANet, FAN,  S$^3$FD and \etc.
 Our DF$^2$S$^2$ is trained on WIDER FACE  training set with data augmentation mentioned in Section \ref{sec:aug},
 and tested on both validation and testing set with multi-scale of $\{600, 800, 1000, 1200, 1400\}$.
Figure \ref{fig:val}  and Figure \ref{fig:test} show the precision-recall curves on WIDER FACE evaluation and testing sets, and Table \ref{table:widerface} summarizes the state-of-the-art results on the WIDER FACE validation set.
Our algorithm obtains the best result in hard subset and competitive results in medium and easy subsets,
\ie 0.956 (Easy), 0.947 (Medium) and 0.898 (Hard) for validation set, and  0.949 (Easy), 0.940 (Medium) and 0.891 (Hard) for testing set.
Considering the hard subset which contains a lot of occluded faces, tiny faces and blurry faces,
Our model outperforms the previous state-art-results of PyramidBox with large margin, $+0.9\%$ in hard task, with the backbone of ResNet50,
which validates the effectiveness of our algorithm in handling high scale variances and occlusion issues.
When using ResNet101 as the backbone, we obtain a great improvements, which proves that our model has many potentials to excavate.

\subsection{Inference Speed}
Our DF$^2$S$^2$ detector is a single-shot detector and thus enjoys high inference speed.
 It runs in real-time inference speed with 26.45 FPS for images of $640 \times 512$ input  size on a computing environment with NVIDIA GPU Tesla P40 and CuDNN-v7.

\section{Conclusion}
This paper proposed a novel framework of 
DF$^2$S$^2$ (Detection with Feature Fusion and Segmentation Supervision) for face detection. 
Our model achieves the state-of-the-art performance on WIDER FACE dataset, yet still enjoys real-time inference speed on GPU due to the nature of the single-stage detection framework.
We present an effective feature fusion pyramids structure and an efficient segmentation branch, both to make model learn better features.
Feature fusion pyramids structure applies semantic information from higher-level feature maps as contextual cues 
to augment low-level feature maps without loss of detailed information in a spatial and channel-wise attention style, making semantics and details complement each other.
And the semantic segmentation branch utilizes detection supervision information to direct models to learn more discriminative features from face regions without comprosing the inference speed.
We note that both of the mentioned ideas are not restricted to face detection tasks, and might also be beneficial to the general object detection task and even the image segmentation task.
For future work, we will  mine more potentials of these two ideas.

{\bf Acknowledgments.} This work was sponsored by DiDi GAIA Research Collaboration Initiative, and partially supported by the National Natural Science Foundation of China under Grant Nos. 61573068 and 61871052, Beijing Nova Program under Grant No. Z161100004916088. We wish to thank Dr. Shifeng Zhang for many helpful discussions.

{\small
\bibliographystyle{ieee}
\bibliography{ms}

\begin{thebibliography}{10}\itemsep=-1pt

\bibitem{DBLP:journals/corr/ChenHW016}
D.~Chen, G.~Hua, F.~Wen, and J.~Sun.
\newblock Supervised transformer network for efficient face detection.
\newblock {\em CoRR}, abs/1607.05477, 2016.

\bibitem{DBLP:journals/corr/ChenWCGXN15}
K.~Chen, J.~Wang, L.~Chen, H.~Gao, W.~Xu, and R.~Nevatia.
\newblock {ABC-CNN:} an attention based convolutional neural network for visual
  question answering.
\newblock {\em CoRR}, abs/1511.05960, 2015.

\bibitem{DBLP:journals/corr/ChenZXNSC16}
L.~Chen, H.~Zhang, J.~Xiao, L.~Nie, J.~Shao, and T.~Chua.
\newblock {SCA-CNN:} spatial and channel-wise attention in convolutional
  networks for image captioning.
\newblock {\em CoRR}, abs/1611.05594, 2016.

\bibitem{DBLP:conf/cvpr/ChenZXNSLC17}
L.~Chen, H.~Zhang, J.~Xiao, L.~Nie, J.~Shao, W.~Liu, and T.~Chua.
\newblock {SCA-CNN:} spatial and channel-wise attention in convolutional
  networks for image captioning.
\newblock In {\em 2017 {IEEE} Conference on Computer Vision and Pattern
  Recognition, {CVPR} 2017, Honolulu, HI, USA, July 21-26, 2017}, pages
  6298--6306, 2017.

\bibitem{DBLP:journals/corr/abs-1809-02693}
C.~Chi, S.~Zhang, J.~Xing, Z.~Lei, S.~Z. Li, and X.~Zou.
\newblock Selective refinement network for high performance face detection.
\newblock {\em CoRR}, abs/1809.02693, 2018.

\bibitem{Corbetta2002}
M.~Corbetta and G.~L. Shulman.
\newblock Control of goal-directed and stimulus-driven attention in the brain.
\newblock {\em Nature Reviews Neuroscience}, 3:201 EP --, Mar 2002.
\newblock Review Article.

\bibitem{DBLP:journals/corr/DaiLHS16}
J.~Dai, Y.~Li, K.~He, and J.~Sun.
\newblock {R-FCN:} object detection via region-based fully convolutional
  networks.
\newblock {\em CoRR}, abs/1605.06409, 2016.

\bibitem{DBLP:conf/iccv/GidarisK15}
S.~Gidaris and N.~Komodakis.
\newblock Object detection via a multi-region and semantic segmentation-aware
  {CNN} model.
\newblock In {\em 2015 {IEEE} International Conference on Computer Vision,
  {ICCV} 2015, Santiago, Chile, December 7-13, 2015}, pages 1134--1142, 2015.

\bibitem{Detectron2018}
R.~Girshick, I.~Radosavovic, G.~Gkioxari, P.~Doll\'{a}r, and K.~He.
\newblock Detectron.
\newblock \url{https://github.com/facebookresearch/detectron}, 2018.

\bibitem{DBLP:journals/corr/HeGDG17}
K.~He, G.~Gkioxari, P.~Doll{\'{a}}r, and R.~B. Girshick.
\newblock Mask {R-CNN}.
\newblock {\em CoRR}, abs/1703.06870, 2017.

\bibitem{DBLP:journals/corr/HeZRS15}
K.~He, X.~Zhang, S.~Ren, and J.~Sun.
\newblock Deep residual learning for image recognition.
\newblock {\em CoRR}, abs/1512.03385, 2015.

\bibitem{DBLP:journals/corr/abs-1709-01507}
J.~Hu, L.~Shen, and G.~Sun.
\newblock Squeeze-and-excitation networks.
\newblock {\em CoRR}, abs/1709.01507, 2017.

\bibitem{DBLP:journals/corr/HuR16}
P.~Hu and D.~Ramanan.
\newblock Finding tiny faces.
\newblock {\em CoRR}, abs/1612.04402, 2016.

\bibitem{DBLP:journals/corr/HuangYDY15}
L.~Huang, Y.~Yang, Y.~Deng, and Y.~Yu.
\newblock Densebox: Unifying landmark localization with end to end object
  detection.
\newblock {\em CoRR}, abs/1509.04874, 2015.

\bibitem{DBLP:conf/nips/KrizhevskySH12}
A.~Krizhevsky, I.~Sutskever, and G.~E. Hinton.
\newblock Imagenet classification with deep convolutional neural networks.
\newblock In {\em Advances in Neural Information Processing Systems 25: 26th
  Annual Conference on Neural Information Processing Systems 2012. Proceedings
  of a meeting held December 3-6, 2012, Lake Tahoe, Nevada, United States.},
  pages 1106--1114, 2012.

\bibitem{DBLP:journals/corr/LinDGHHB16}
T.~Lin, P.~Doll{\'{a}}r, R.~B. Girshick, K.~He, B.~Hariharan, and S.~J.
  Belongie.
\newblock Feature pyramid networks for object detection.
\newblock {\em CoRR}, abs/1612.03144, 2016.

\bibitem{DBLP:journals/corr/abs-1708-02002}
T.~Lin, P.~Goyal, R.~B. Girshick, K.~He, and P.~Doll{\'{a}}r.
\newblock Focal loss for dense object detection.
\newblock {\em CoRR}, abs/1708.02002, 2017.

\bibitem{DBLP:journals/corr/LiuAESR15}
W.~Liu, D.~Anguelov, D.~Erhan, C.~Szegedy, S.~E. Reed, C.~Fu, and A.~C. Berg.
\newblock {SSD:} single shot multibox detector.
\newblock {\em CoRR}, abs/1512.02325, 2015.

\bibitem{DBLP:journals/corr/LongSD14}
J.~Long, E.~Shelhamer, and T.~Darrell.
\newblock Fully convolutional networks for semantic segmentation.
\newblock {\em CoRR}, abs/1411.4038, 2014.

\bibitem{DBLP:journals/corr/LuoLUZ17}
W.~Luo, Y.~Li, R.~Urtasun, and R.~S. Zemel.
\newblock Understanding the effective receptive field in deep convolutional
  neural networks.
\newblock {\em CoRR}, abs/1701.04128, 2017.

\bibitem{DBLP:journals/corr/MnihHGK14}
V.~Mnih, N.~Heess, A.~Graves, and K.~Kavukcuoglu.
\newblock Recurrent models of visual attention.
\newblock {\em CoRR}, abs/1406.6247, 2014.

\bibitem{DBLP:journals/corr/abs-1708-03979}
M.~Najibi, P.~Samangouei, R.~Chellappa, and L.~S. Davis.
\newblock {SSH:} single stage headless face detector.
\newblock {\em CoRR}, abs/1708.03979, 2017.

\bibitem{DBLP:journals/corr/Ohn-BarT17}
E.~Ohn{-}Bar and M.~M. Trivedi.
\newblock To boost or not to boost? on the limits of boosted trees for object
  detection.
\newblock {\em CoRR}, abs/1701.01692, 2017.

\bibitem{DBLP:journals/corr/RedmonDGF15}
J.~Redmon, S.~K. Divvala, R.~B. Girshick, and A.~Farhadi.
\newblock You only look once: Unified, real-time object detection.
\newblock {\em CoRR}, abs/1506.02640, 2015.

\bibitem{DBLP:journals/corr/RenHG015}
S.~Ren, K.~He, R.~B. Girshick, and J.~Sun.
\newblock Faster {R-CNN:} towards real-time object detection with region
  proposal networks.
\newblock {\em CoRR}, abs/1506.01497, 2015.

\bibitem{Song_2018_CVPR}
C.~Song, Y.~Huang, W.~Ouyang, and L.~Wang.
\newblock Mask-guided contrastive attention model for person re-identification.
\newblock In {\em The IEEE Conference on Computer Vision and Pattern
  Recognition (CVPR)}, June 2018.

\bibitem{DBLP:conf/cvpr/TaigmanYRW14}
Y.~Taigman, M.~Yang, M.~Ranzato, and L.~Wolf.
\newblock Deepface: Closing the gap to human-level performance in face
  verification.
\newblock In {\em 2014 {IEEE} Conference on Computer Vision and Pattern
  Recognition, {CVPR} 2014, Columbus, OH, USA, June 23-28, 2014}, pages
  1701--1708, 2014.

\bibitem{DBLP:journals/corr/abs-1803-07737}
X.~Tang, D.~K. Du, Z.~He, and J.~Liu.
\newblock Pyramidbox: {A} context-assisted single shot face detector.
\newblock {\em CoRR}, abs/1803.07737, 2018.

\bibitem{DBLP:conf/iccv/ViolaJ01}
P.~A. Viola and M.~J. Jones.
\newblock Robust real-time face detection.
\newblock In {\em {ICCV}}, page 747, 2001.

\bibitem{DBLP:journals/corr/WangLJW17}
H.~Wang, Z.~Li, X.~Ji, and Y.~Wang.
\newblock Face {R-CNN}.
\newblock {\em CoRR}, abs/1706.01061, 2017.

\bibitem{DBLP:journals/corr/abs-1711-07246}
J.~Wang, Y.~Yuan, and G.~Yu.
\newblock Face attention network: An effective face detector for the occluded
  faces.
\newblock {\em CoRR}, abs/1711.07246, 2017.

\bibitem{DBLP:journals/corr/XuBKCCSZB15}
K.~Xu, J.~Ba, R.~Kiros, K.~Cho, A.~C. Courville, R.~Salakhutdinov, R.~S. Zemel,
  and Y.~Bengio.
\newblock Show, attend and tell: Neural image caption generation with visual
  attention.
\newblock {\em CoRR}, abs/1502.03044, 2015.

\bibitem{DBLP:journals/corr/YangLLT15b}
S.~Yang, P.~Luo, C.~C. Loy, and X.~Tang.
\newblock {WIDER} {FACE:} {A} face detection benchmark.
\newblock {\em CoRR}, abs/1511.06523, 2015.

\bibitem{DBLP:journals/corr/YangXLT17}
S.~Yang, Y.~Xiong, C.~C. Loy, and X.~Tang.
\newblock Face detection through scale-friendly deep convolutional networks.
\newblock {\em CoRR}, abs/1706.02863, 2017.

\bibitem{DBLP:journals/corr/YuJWCH16}
J.~Yu, Y.~Jiang, Z.~Wang, Z.~Cao, and T.~S. Huang.
\newblock Unitbox: An advanced object detection network.
\newblock {\em CoRR}, abs/1608.01471, 2016.

\bibitem{DBLP:journals/corr/ZeilerF13}
M.~D. Zeiler and R.~Fergus.
\newblock Visualizing and understanding convolutional networks.
\newblock {\em CoRR}, abs/1311.2901, 2013.

\bibitem{DBLP:conf/cvpr/ZeilerKTF10}
M.~D. Zeiler, D.~Krishnan, G.~W. Taylor, and R.~Fergus.
\newblock Deconvolutional networks.
\newblock In {\em The Twenty-Third {IEEE} Conference on Computer Vision and
  Pattern Recognition, {CVPR} 2010, San Francisco, CA, USA, 13-18 June 2010},
  pages 2528--2535, 2010.

\bibitem{DBLP:journals/corr/abs-1712-00721}
J.~Zhang, X.~Wu, J.~Zhu, and S.~C.~H. Hoi.
\newblock Feature agglomeration networks for single stage face detection.
\newblock {\em CoRR}, abs/1712.00721, 2017.

\bibitem{DBLP:journals/corr/ZhangZL016}
K.~Zhang, Z.~Zhang, Z.~Li, and Y.~Qiao.
\newblock Joint face detection and alignment using multi-task cascaded
  convolutional networks.
\newblock {\em CoRR}, abs/1604.02878, 2016.

\bibitem{DBLP:conf/cvpr/ZhangWBLL18}
S.~Zhang, L.~Wen, X.~Bian, Z.~Lei, and S.~Z. Li.
\newblock Single-shot refinement neural network for object detection.
\newblock In {\em 2018 {IEEE} Conference on Computer Vision and Pattern
  Recognition, {CVPR} 2018, Salt Lake City, UT, USA, June 18-22, 2018}, pages
  4203--4212, 2018.

\bibitem{zhang2019single-shot}
S.~Zhang, L.~Wen, H.~Shi, Z.~Lei, S.~Lyu, and S.~Z. Li.
\newblock Single-shot scale-aware network for real-time face detection.
\newblock {\em International Journal of Computer Vision}, pages 1--23, 2 2019.

\bibitem{DBLP:journals/corr/abs-1901-06651}
S.~Zhang, R.~Zhu, X.~Wang, H.~Shi, T.~Fu, S.~Wang, T.~Mei, and S.~Z. Li.
\newblock Improved selective refinement network for face detection.
\newblock {\em CoRR}, abs/1901.06651, 2019.

\bibitem{DBLP:conf/icb/ZhangZLSWL17}
S.~Zhang, X.~Zhu, Z.~Lei, H.~Shi, X.~Wang, and S.~Z. Li.
\newblock Faceboxes: {A} {CPU} real-time face detector with high accuracy.
\newblock In {\em 2017 {IEEE} International Joint Conference on Biometrics,
  {IJCB} 2017, Denver, CO, USA, October 1-4, 2017}, pages 1--9, 2017.

\bibitem{DBLP:journals/corr/abs-1708-05237}
S.~Zhang, X.~Zhu, Z.~Lei, H.~Shi, X.~Wang, and S.~Z. Li.
\newblock S\({}^{\mbox{3}}\)fd: Single shot scale-invariant face detector.
\newblock {\em CoRR}, abs/1708.05237, 2017.

\bibitem{DBLP:journals/ijon/ZhangZLWSL18}
S.~Zhang, X.~Zhu, Z.~Lei, X.~Wang, H.~Shi, and S.~Z. Li.
\newblock Detecting face with densely connected face proposal network.
\newblock {\em Neurocomputing}, 284:119--127, 2018.

\bibitem{DBLP:journals/corr/abs-1712-00433}
Z.~Zhang, S.~Qiao, C.~Xie, W.~Shen, B.~Wang, and A.~L. Yuille.
\newblock Single-shot object detection with enriched semantics.
\newblock {\em CoRR}, abs/1712.00433, 2017.

\bibitem{DBLP:journals/corr/abs-1802-09058}
C.~Zhu, R.~Tao, K.~Luu, and M.~Savvides.
\newblock Seeing small faces from robust anchor's perspective.
\newblock {\em CoRR}, abs/1802.09058, 2018.

\bibitem{DBLP:journals/corr/ZhuZLS16}
C.~Zhu, Y.~Zheng, K.~Luu, and M.~Savvides.
\newblock {CMS-RCNN:} contextual multi-scale region-based {CNN} for
  unconstrained face detection.
\newblock {\em CoRR}, abs/1606.05413, 2016.

\bibitem{DBLP:journals/corr/ZhuLLSL15}
X.~Zhu, Z.~Lei, X.~Liu, H.~Shi, and S.~Z. Li.
\newblock Face alignment across large poses: {A} 3d solution.
\newblock {\em CoRR}, abs/1511.07212, 2015.

\bibitem{DBLP:conf/cvpr/ZhuLYYL15}
X.~Zhu, Z.~Lei, J.~Yan, D.~Yi, and S.~Z. Li.
\newblock High-fidelity pose and expression normalization for face recognition
  in the wild.
\newblock In {\em {IEEE} Conference on Computer Vision and Pattern Recognition,
  {CVPR} 2015, Boston, MA, USA, June 7-12, 2015}, pages 787--796, 2015.

\end{thebibliography}
}

\end{document}